\documentclass[sn-mathphys-num]{sn-jnl}

\usepackage{graphicx}%
\usepackage{amsmath}%
\usepackage{orcidlink}

\begin{document}

\title[The Return of Pseudosciences in AI: Have ML and DL Forgotten Lessons from Statistics and History?]{The Return of Pseudosciences in Artificial Intelligence: Have Machine Learning and Deep Learning Forgotten Lessons from Statistics and History?}


\author[1]{Pr. \fnm{Jérémie} \sur{Sublime} \orcidlink{0000-0003-0508-8550}}\email{jsublime@isep.fr}

\affil[1]{\orgname{ISEP - Paris Institute of Digital Technologies}, \orgaddress{\street{10 rue de Vanves}, \city{Issy-Les-Moulineaux}, \postcode{92130}, \country{France}}}


\abstract{In today’s world, AI programs powered by Machine Learning are ubiquitous, and have achieved seemingly exceptional performance across a broad range of tasks, from medical diagnosis and credit rating in banking, to theft detection via video analysis, and even predicting political or sexual orientation from facial images. These predominantly deep learning methods excel due to their extraordinary capacity to process vast amounts of complex data to extract complex correlations and relationship from different levels of features.

In this paper, we contend that the designers and final users of these ML methods have forgotten a fundamental lesson from statistics: correlation does not imply causation. Not only do most state-of-the-art methods neglect this crucial principle, but by doing so they often produce nonsensical or flawed causal models, akin to social astrology or physiognomy. Consequently, we argue that current efforts to make AI models more ethical by merely reducing biases in the training data are insufficient. Through examples, we will demonstrate that the potential for harm posed by these methods can only be mitigated by a complete rethinking of their core models, improved quality assessment metrics and policies, and by maintaining humans oversight throughout the process.}




\maketitle

\section{Introduction}
\label{Introduction}

Since the advent of Deep Learning technologies in recent years, artificial intelligence (AI) systems have become pervasive across a wide range of real-wold applications. Indeed, Machine Learning (ML) specialists have been busy generating a continuous flow of AI solutions targeted at all sorts of applications which are often far beyond their own areas of expertise: theft detection via video surveillance  in local shopping centres, credit risk analysis software in banking, marketing algorithms that determine which advertisements to display based on personal data,  AI assisted diagnosis, etc. There are more controversial examples, such as seemingly innocuous facial analysis software designed to infer political or orientation, and even AI-driven judicial decision to decide  to decide bail eligibility based on an individual’s background data. The list of applications, some trivial and others deeply consequential, is ever-expanding with each passing day.

Given the seriousness and sensitivity of some of these tasks delegated to AI systems, there is a growing interest within both the ML and ethics communities to advocate for fairer and socially responsible AI algorithms \cite{weinberg2022rethinking,cheng2021socially} that minimize harm's risks and are as unbiased as possible. This movement has taken various forms including the development of explainable AI \cite{ras2022explainable}, and a stronger focus on training Machine Learning algorithms using data that have been curated to reduce biases of all kinds \cite{caton2022impact}.

However, it is this paper's goal to demonstrate that possible biases in data  represent only a small part of the issue, and that the primary ethical concern with current machine learning and deep learning methods lies in the undue attribution of causality by their designers and users. Indeed, while it is undeniable that Deep Learning methods -and in particular convolutional networks \cite{7298594} for images and videos, as well as autoencoders \cite{NIPS1993_9e3cfc48} for complex data- are highly effective at identifying complex and intricate relationships as well as correlations from large amount of training data, it remains a fundamental error to assume that such systems, which are inherently statistical, can be trusted for sensitive tasks that should require explainability. We contend that bestowing these deep learning-based systems with what amounts to "oracle-like" powers is not only selling snake oil, but also akin to endorsing pseudosciences such as Lombrosianism, physiognomy, and social astrology. Moreover, we argue that, in addition to their lack of knowledge and disregard for historical context in various application domains, too many Machine Learning researchers seem have forgotten the fact that the field of machine learning originated as a branch of statistics, where a key tenet is that correlation does not imply causation.

Finally, we assert that efforts at developing AI algorithms that ``do no harm" are destined to fail as long as AI projects formerly reliant on human intervention are driven exclusively by ML experts  who are too often removed from the application domain, and who rely on metrics that prioritise outcomes over fairness. 

Within this context, this paper will develop the following points:
\begin{itemize}
    \item Through an exploration of various recent machine learning applications, either currently in use or nearing deployment, we will demonstrate how pseudosciences such as physiognomy and racial theories have been revived, rebranded with a modern veneer, and even legitimised through the use of AI.
    \item By examining the metrics used to validate these methods and exploring simple probabilities, we will highlight the substantial harm these algorithms can cause when applied in critical public sectors such as justice and security.
    \item We will then address how the prevailing focus on reducing bias through curated training data, while promoting AI fairness, fails to tackle the core issue, which lies in the models themselves.  Indeed, the ``theory-free" argument put forward by proponents of current AI methods not only makes biases more challenging to detect, but is frequently used alongside high-quality metrics as a misleading justification for the alleged fairness of a system.
    \item Finally, we will offer recommendations on how some of these issues should be addressed: prioritising metrics that promote fairness over mere performance; abandoning the idea that AI systems can be used as oracles and could replace decades of expertise that ML researchers and AI algorithms sorely lack; and reminding ML researchers and AI users that, even when prompted by a high performing very deep neural networks, patterns and correlations still do not imply causation. Believing otherwise invites disaster, or —as argued in this paper— the resurgence of pseudoscientific beliefs we presumed had been discarded nearly 80 years ago.
\end{itemize}

\section{State of the Art on potentially misguided and harmful AI applications}

In this section, we provide a comprehensive overview of recent AI applications that have a strong potential to be harmful, or could at the very least be misguided. To clarify our use of the terms \emph{potentially harmful} and \emph{misguided}, we will focus on those AI applications that could lead to critical, life-altering decisions or have a significant impact on people's safety and privacy. These applications can broadly be divided into two categories, which may at time overlap: those related to justice, security and law enforcement; and applications with sociological or societal implications, including the health sector.

In the field of justice and law enforcement, several AI systems have been tested already, and attempts to forecast criminal behaviour through AI and statistical methods are not particularly new \cite{berk2009forecasting,tollenaar2013method}. However, the deployment of AI systems from machine learning research labs into actual courtrooms is a more recent development. An example of this is the OASys (Offender Assessment System) AI in the United Kingdom \cite{Oasys,doi:10.1177/0093854811431239} which assists probation officers in assessing the risk posed by individuals within the justice system and determining the best course of action. Similar systems for criminal risk assessment have been explored for suitability in Thailand \cite{butsara2019predicting}, the USA \cite{duwe2017out,ozkan2020predicting}, Finland \cite{salo2019predictive}, the Netherlands \cite{tollenaar2013method,tollenaar2019optimizing}, and there is a substantial body of literature discussing such AI programs at various stages of development and implementation  \cite{travaini2022machine}. Although few of these systems have been fully integrated thus far, the reduced costs compared with human operators, along with bold claims of high accuracy, suggest that this trend will continue to grow. 

In fact, AI has already become prevalent in one area of security: video surveillance. In this area, numerous ssystems have been developed, not only for facial recognition \cite{zalnieriute2020burning}, but also to automatically detect thefts in banks \cite{kakadiya2019ai}, tracking thieves via CCTV and tracking \cite{varun2023real}, or identifying fraud and deception during online exams and job interviews \cite{radwan2022one,el2022deep,tsuchiya2023detecting}. 

While more efficient justice systems, increased security, faster processes, and improved chances of apprehending thieves and fraudsters are undoubtedly commendable goals, serious concerns arise when AI begins to replace human judgment and makes decisions based on factors such as skin colour, social background, religion, real or perceived sexual orientation, or other obscure and frequently misbegotten criteria. Although DL and AI technologies often hide these criteria behind their deep layers, some research is explicit enough to understand that AI specialists may well try to sell systems that allegedly could detect future criminality based on facial features and gestural tics \cite{wu2016automated,wu2017responsescritiquesmachinelearning,kabir2020human,hashemi2020retracted,verma2020convolutional}.

Other systemic uses of AI with a sociological or societal impact are just as numerous as those mentioned earlier in law enforcement, and often revolve around profiling individuals for various purposes. 
Marketing is perhaps the most well-known of these purposes, and also the most innocuous: such AI programs are referred to as recommender systems \cite{doi:10.1080/01621459.2023.2279695}. These systems aim to target the right advertisements to the right individual, increasing the likelihood of converting views into sales. AI sponsored advertising selects online ads, is embedded into smart TVs, and is also used by retailers to recommend products based on prior purchases.
However, other categories of AI-based profiling carry more serious implications. One example is credit rating algorithms, which determine who is eligible for a loan, whether it be a small loan for e-commerce \cite{sahoo2023faulty}, or a larger one for purchasing a home \cite{chun2021study}. Furthermore, despite several regulations such as in the European Union, the explainability of these models remains challenging and costly \cite{dessain2023cost}, and they have often proven to perpetuate the same racial or social-economic biases as other AI methods.

Finally, we can highlight a range of AI methods that claim to detect various traits with high accuracy based solely on facial features from static images. These include detecting sexual orientation \cite{wang2018deep}, personality traits \cite{kachur2020assessing,parde2019social,peterson2022deep}, and even so-called ``\emph{human abnormality}" \cite{kabir2020human} -a term used by the authors to encompass conditions ranging from mental illness and personality disorders to autism and criminality-.  Additionally, there are claims of detecting autism \cite{alam2022empirical,mujeeb2022identification} and political orientation \cite{kosinski2024facial} using similar DL methods. Despite many of them looking like pseudo-scientific quackery, several of such methods have been commercialized, as seen in products like \emph{Faception} \footnote{https://www.faception.com/} or \emph{Hirevue} \footnote{https://www.hirevue.com/}, which can be used by private companies to make hiring decisions. It is not difficult to imagine how such seemingly far-fetched neural networks could be integrated with face recognition systems or other AI tools in law enforcement, raising significant concern about the potential harm they may cause. Furthermore, while Western countries tend to have stronger regulations aimed at preventing blatant discrimination by such systems, these regulations are not universal, and even when such regulations exist, proving discrimination by an AI system remains a substantial challenge. As for countries countries with fewer regulations, such systems could easily lead to systemic surveillance and discrimination against minority groups.

\section{The reanimation of pseudosciences by ML methods and its ethical implications}

\subsection{Basics of Deep Learning and ML inference}
\label{sec:DL}

To begin explaining how the Machine Learning and Deep Learning applications we have just discussed have revived pseudoscientific ideas, we must first explore the basic functionalities of these methods and the assumptions they rely on. In other words, we will first clarify how these methods learn and operate.

Within the context of this papers, we focus on two specific supervised Machine Learning tasks: 
\begin{itemize}
    \item Classification: It consists in accurately predicting a category, a class or a label based on the attributes of observed data.
    \item Regression: A similar task, but where the goal is to predict the values (numerical value) of one or more variables based on other attributes.
\end{itemize}

To accomplish these tasks, Machine Learning methods use evidence, referred to as training data, to perform predictions or classifications. The aim is therefore to generalise from the training data (also called the training set) to new, and previously unseen data (the test set). It can therefore be asserted that machine learning is fundamentally about inductive inference.

Machine Learning methods are built on a variety of principles, primarily derived from statistical and probabilistic models. However, it is the advances in Deep Learning and neural networks technologies  that have propelled the field forward, enabling the analysis of data types that were previously too challenging for traditional ML methods, such as images, videos, time series, and even conventional data with a very large number of features.

\begin{figure}[!h]
    \centering
    \includegraphics[width=0.9\textwidth]{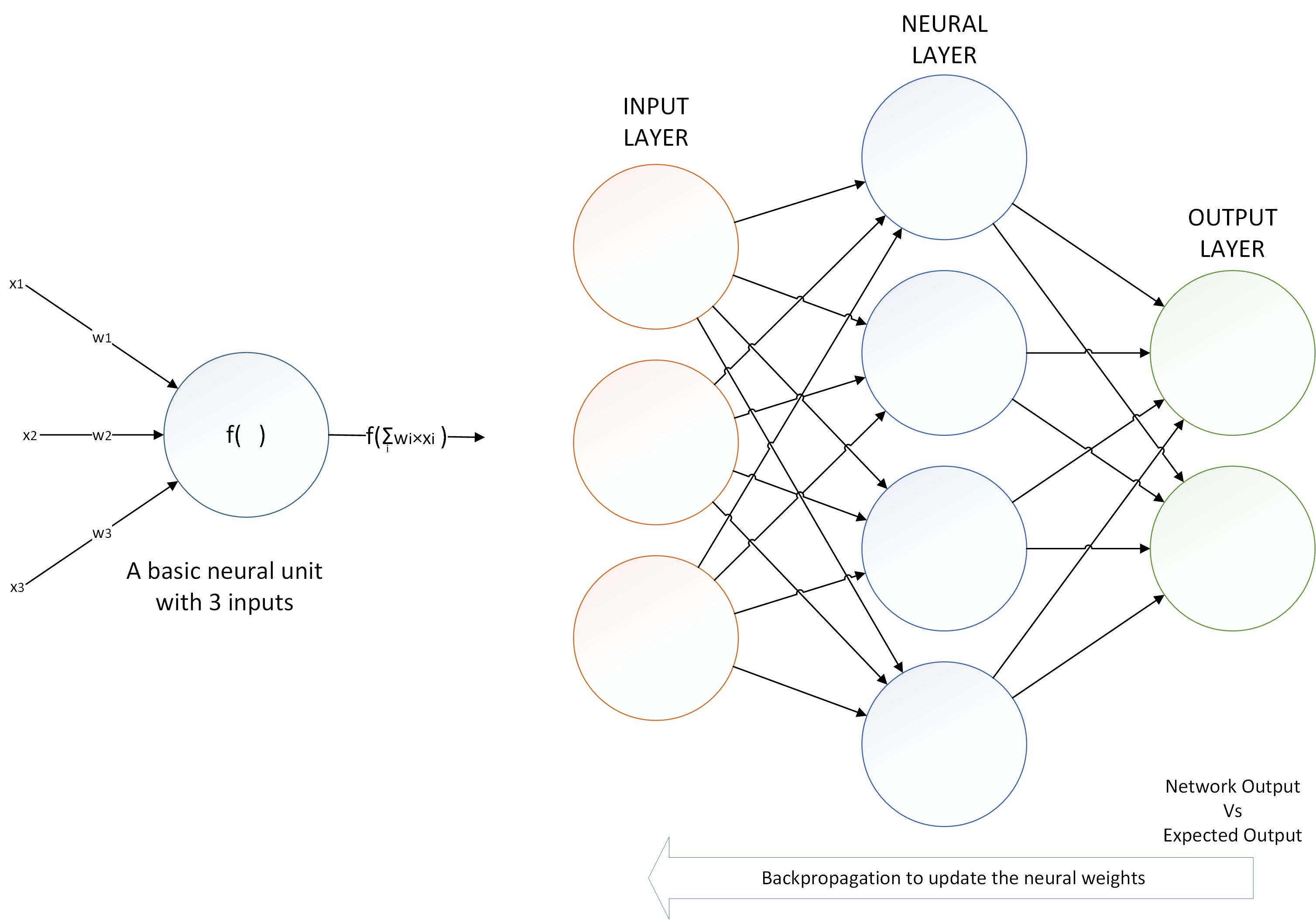}
    \caption{On the left: a basic neural unit with 3 weighted inputs. - On the right: a simple network with a 3 features input layer, a 2-class output layer and a single neural layer in the middle. -- This figure shows how complex linear combinations of the original input features can be computed using different layers with activation functions $f(\cdot)$.}
    \label{fig:basic-NN}
\end{figure}

A neural network is a complex algorithm that takes a large number of features as inputs: a large number of variables, pixels or patches of pixels from an image, etc. The network then computes intermediate representations within multiple layers \cite{erhan2009visualizing}, and produces an output which can be a class, a label, or a value, depending on the task. An example of a simple neural network is shown in Figure \ref{fig:basic-NN}, and some of the most commonly used activation functions are listed in Table \ref{tab:activation}. Training such a network involves presenting it with a large set of labelled examples to tune its weight parameters. This process is known as parametrisation and relies on gradient backpropagation with respect to some objective function, which typically seeks to minimise the error between the network's output and the expected result from the training set label.
Through this training process, the network's weight will converge, transforming said network into a highly complex mathematical function that maps inputs to outputs. From there, assuming the relevant hypotheses of similarity between the training and the test sets hold, the network should be able to infer outcomes for previously unseen instances, and to generalise from particular cases to broader classes \cite{mohri2018foundations,jo2021machine}.

\begin{table}[!h]
\begin{tabular}{|c|c|c|}
 \hline
\textbf{Hyperbolic Tangent} & \textbf{Logistic} & \textbf{Rectified Linear Unit (ReLU)} \\
\hline
$tanh(x)=\frac{e^x - e^{-x}}{e^x + e^{-x}}$ & $\sigma(x)=\frac{1}{1+e^{-x}}$ & $ReLU(x)= max(0,x)$  \\
\hline
 \textbf{Softplus} & \textbf{Sigmoid} & \textbf{Softmax} (vector output) \\
\hline
$ln(1+e^x)$ & $\frac{x}{1+e^{-x}}$ & $ \forall  i \quad soft_i=\frac{e^{x_i}}{\sum_{j=1}^J e^{x_j}}$ \\
\hline
\end{tabular}
\caption{Example of common activation functions: The $x_i$ are the inputs, and we have $x=\sum_i w_i x_i$, with $w_i$ the neural weights.}
\label{tab:activation}
\end{table}

While we do not dispute the reality of correlations identified by Deep Learning algorithms, nor their high efficiency, or ability to generalize -though we do not exclude possible bias in the datasets, which we will discuss in a later section-, we raise concern about two specific aspects: 
\begin{itemize}
    \item The way Deep Learning methods appears to implicitly infer causation from correlations and patterns as their primary mode of induction. Indeed, and while not specific to Deep Learning, input variables are de facto causal variables for Machine Learning Models to make their decisions \cite{bengio2013representation,bengio2019meta}.
    \item The validity of the interpretation lent to said patterns and correlations, whenever interpretation is even possible at all. 
\end{itemize} 

As these two concerns are closely intertwined, we will use an example to illustrate why we believe deep learning faces challenges in its approach to handling causality.

Let us consider the field of medicine, a domain we have not yet discussed but which has also seen substantial investment in AI methods recently, particularly for tasks such as automated diagnosis. Anyone with even a basic scientific understanding knows that diseases cause symptoms; causation flows from the disease to the symptoms. When a medical professional tries to diagnose a patient, they ask questions and order tests iteratively to confirm or rule out possible conditions, based on the knowledge of which disease causes which symptoms. Each new piece of information -from a test or a patient answer- helps narrowing down potential diseases, and may lead to further questions or tests becoming relevant or irrelevant. The investigation then continues until the disease is correctly identified. This process resembles the game ``Guess Who?". This is not how AI methods based on Deep Learning approach automated diagnosis. First, these systems lack any field-specific or expert knowledge, meaning they do not recognise the fundamental mechanism that diseases cause symptoms. Second, because these methods are trained on large datasets containing patient information, test results, symptoms, and associated diagnoses, the direction of causation is effectively reversed: it is the symptoms (often complex combinations of them) that determine the disease, rather than the disease causing the symptoms. While medical professionals iteratively process symptoms and test results to find out the underlying condition based on mechanisms,  based on causal mechanisms to identify the condition, AI methods process all possible symptoms and tests regardless of their relevance to produce a diagnosis that best matches what they have encountered in their training data.

We acknowledge that AI’s comprehensive approach, which includes all symptoms and their complex interactions, can sometimes offer advantages over clinical expertise by analysing parameters that a clinician might not have considered. However, our point is that the AI approach disregards underlying mechanisms and builds causality in unconventional ways.


While this example illustrate a peculiar form of backward causation in the field of medicine, the same issue exist in all applications of Deep Learning. Indeed, because deep learning fundamentally involves constructing complex mathematical functions to identify correlations and patterns from features, and then using these to infer classifications or regressions, it is susceptible to the same types of unintended, backward, and unfounded implicit causation mechanisms across various applications of deep learning algorithms.

\subsection{The silent return of physiognomy, Lombrosianism, phrenology, distorted sociobiology, social astrology and other quackeries with a new AI polish}

Unless you are a native english speaker with a background in ethics, philosophy or modern history, you may not be familiar with terms such as \emph{physiognomy}, \emph{phrenology} and \emph{Lombrosianism}. However, you may have heard about sociobiology and social astrology. Before we move forward with our discussion on the susceptibility of machine learning and deep learning to pseudoscience, let us define some of these terms:

\paragraph{
\textit{Physiognomy is ``the facility to identify, from the form and constitution of external parts of the human body, chiefly the face, exclusive of all temporary signs of emotions, the constitution of the mind and the heart." -- Georg Christoph Lichtenberg, 1778}}

However, physiognomy is much older than Lichtenberg, and already 2 millenia before dawn, the Babylonians spoke of \emph{``physiognomic omens"} regarding facial features that could be predictive of life trajectory \cite{jones2018cambridge}.

\paragraph{
\textit{Phrenology -or craniology- involves the measurement of bumps on the skull to predict mental traits.}} Originally proposed by German physician Franz Joseph Gall in 1796 when he was trying to figure out the function of different brain areas. The discipline became popular in the 1830s and 1840s -in particular in the USA- where it was used as an argument by physicians such as Charles Caldwell in an attempt to prove the superiority of white people over African people (thus justifying slavery and segregation), and by Samuel Morton to justify the persecution of native Americans.

\paragraph{\textit{Lombrosianism is a theory in criminology developed in the late 19th century by Italian physician Cesare Lombroso. This theory suggest that criminal behavior is innate and can be identified through physical traits. Mister Lombroso believed that criminals were biologically different from non-criminals, often marked by ``atavistic" features that resembled earlier stages of human evolution (such as certain facial structures or body types). This theory supports the idea that criminals are ``born," not made, and could be distinguished by these primitive traits.}} \cite{lombroso1891illustrative}.

All three pseudosciences were extensively used for racist purposes, including justification of slavery and imperialism during the 18th and 19th century, and later for eugenics and other imperialist policies in the 20th century under the German Third Reich.

Distorted sociobiology is nothing less than the modern version of Lombrosianism, phrenology and and physiognomy. It seeks to apply and distort advances in biology and genetics to predict similar outcomes (such as life trajectories, IQ, predispositions to criminal behavior, propensity to lie, mental traits, and sexual orientation) based on factors like ethnic background -which often means ``race"-, social group (religious or cultural), genetics, neurochemistery, or behavioral traits (such as heart beat, eye movements and voice patterns). As for social astrology, while seemingly more innocuous, it is akin to these disciplines in its attempt to infer characteristics like political orientation, sexual orientation, or religion based on similar features, along with additional data about individuals' social backgrounds.

\begin{table}[!h]
\begin{tabular}{|p{4cm}|p{8.5cm}|}
\hline
Pseudoscience & References \\
\hline
\hline
Physiognomy & \cite{peterson2022deep}, \cite{wang2018deep}, \cite{kachur2020assessing}, \cite{kabir2020human}, \cite{kosinski2024facial}, \cite{verma2020convolutional}, \cite{parde2019social} \\
\hline
Lombrosianism & \cite{travaini2022machine}, \cite{wu2016automated}, \cite{hashemi2020retracted}, \cite{verma2020convolutional} \\
\hline
Distorted sociobiology & \cite{tsuchiya2023detecting}, \cite{mujeeb2022identification} \\
\hline
Social Astrology & \cite{berk2009forecasting},\cite{doi:10.1177/0093854811431239}, \cite{tollenaar2013method,tollenaar2019optimizing}, \cite{butsara2019predicting}, \cite{duwe2017out}, \cite{ozkan2020predicting}, \cite{norori2021addressing} \\
\hline
\end{tabular}
\caption{ML papers about pseudosciences: Some of them are selling pseudosciences, other are review papers or critics of pseudoscientific practices. Some papers may cover more than one category.}
\label{tab:categories}
\end{table}

Now that you are more familiar with the definitions of these pseudosciences, it should be clearer that many of the AI algorithms discussed so far in this work are part of these pseudo-disciplines. Table \ref{tab:categories} summarises the references we have examined, organised in different categories. Please note that papers listed in this table are by no mean all pseudoscientific -though some are-; this table also includes review papers and critiques of ML applications within these fields. Furthermore, Artificial Intelligence is not the only field that during a rapid expansion period has been plagued with a pseudoscience issue as well as confusion between correlations and causality: This was, for instance, also the case for genetics and genomics in the 90s a,d early 2000s \cite{bailey1991genetic,lewontin2006analysis}.

While other researchers have already pointed out the vulnerability of Machine Learning to pseudosciences \cite{andrews2024reanimation}, we contend that not only is Machine Learning vulnerable to pseudosciences, but that it also provides them with a new veneer of legitimacy: Lombrosianism, physiognomy and phrenology were discredited sciences rooted in racist biases. Distorted sociobiology selectively manipulated and twisted robust biology concepts to justify indefensible claims. Social Astrology similarly distorts concepts by adding layers of sociology (rather than biology) and statistics, all the while carefully avoiding the issue of correlation versus causation. And bow, AI represents the latest refinement of these pseudosciences, embedding racism and bigotry so deeply within neural layers that their presence is nearly obscured, and the existence of such AI algorithms being further legitimized by ``high scores on result metrics".

\section{Impact of ML quality metrics on the social harm potential of AI algorithms}

\subsection{ML quality metrics for classification}

Since we have just discussed the legitimization of AI methods in part due to their high scores on quality metrics, in this section we will examine some of these metrics and how they are used to assess the value of Machine Learning and Deep Learning algorithms. We will particularly focus on the quality metrics used for classification tasks, as this type of Machine leanring methods are most susceptible to pseudoscientific distortion.

In essence, the goal of any Machine Learning classification method is to achieve high accuracy on the training and test set to ensure robust generalization features. Accuracy is simply defined as the number of correctly classified instance divided by the total number of instance. While classifier can be multi-class (predicting the correct class or label among more than two categories) or binary (determining whether an observed data belongs to a given single class or not), multi-class and binary classifiers are typically evaluated in the same way, with each multi-class label being assessed individually as if it were binary. Binary quality measures are then computed using the following elements:
\begin{itemize}
    \item Number of true positive cases (TP) : Data assigned to a class and that really belong to this class.
    \item Number of false positive cases (FP): Data assigned to a class but that do not belong to it.
    \item Number of true negative cases (TN): Data not assigned to a class and that really do not belong to this class.
    \item Number of false negative cases (FN): Data not assigned to a class but that should belong to it.
\end{itemize}

Using these notations, the classifier's accuracy for a given class is computed as follows:

\begin{equation}
    Accuracy = \frac{TP+TN}{TP+TN+FP+FN}
\end{equation}

Another very common binary measure is the recall, which in some fields is also called the ``hit rate" or ``sensitivity". It is the percentage of correctly detected positive cases:

\begin{equation}
    Recall = \frac{TP}{TP+FN}
\end{equation}

Since accuracy is the most well known quality index inside and outside the Machine Learning community, it comes to no surprise that this metric is frequently used in applied machine learning papers and in marketing AI products that are intended for public release. Following accuracy, recall is another common secondary quality metric, particularly in security applications where detecting as many positive cases as possible (e.g., identifying all criminals) is often a key performance indicator. This is similarly important in health-related applications where missing a deadly disease is more critical than a false positive diagnosis.

Focusing primarily on these two metrics can pose significant issues regarding the potential harm of AI algorithms. For example, what about the consequences of false positives in criminal conviction ? Or the impact of being unfairly rejected for a job due to a face recognition system making an erroneous assessment ? Notice how we are not even talking about biases here. Rather, we are addressing the inadequacy of using the wrong metrics to evaluate the safety of AI systems. Yet, there are metrics that could be used to mitiate such risks. The precision for instance is the probability that a predicted positive case is really positive:

\begin{equation}
    Precision = \frac{TP}{TP+FP}
    \label{eq:prec}
\end{equation}

There is also its opposite, the specificity (or true negative rate) which measures how well a binary classifier identifies negative cases. Specificity is important when the cost of false positives is high, as it indicates the model's ability to correctly reject negative cases:

\begin{equation}
    Specificity= \frac{TN}{TN+FP}
\end{equation}

Despite the existence of these important quality metrics, accuracy and, to a lesser extent, recall are typically highlighted first and prominently featured in abstracts or when advertising for an AI system. Precision and specificity, on the other hand, are often obscured within metrics such as the Area Under the Curve (AUC) or Receiver Operating Characteristic (ROC) curves: both are types of diagram which assess classifier performances on true positive and true negative rates under different parameters settings. Alternatively, these metrics may be integrated into composite indices like the F-Score. Such approaches make it very challenging to properly assess how well an AI method performs in terms of precision and specificity.

\begin{equation}
    F\text{-}Score= \frac{2 \times precision \times recall}{precision + recall}
\end{equation}

Furthermore, it is evident to us that the practice of systematically obscuring metrics that could reveal the potential harm of AI methods contributes significantly to the proliferation of pseudoscientific approaches within the field of AI.

\subsection{Assessing the real impact of error made by AI systems}

To evaluate the potential impact of these systems if implemented at a large scale, as well as the real impact of quality metrics, we propose using available data and estimates about population and CCTV cameras in four large metropolitan areas: London (UK), Beijing (China), Hyderabad (India) and New York (USA). Indeed, given current surveillance trends, CCTV cameras, which are already extensively used for security purposes in many countries, are likely to be the primary medium for deploying AI systems based on physiognomy or Lombrosianism for crime detection and analysis.

For our simulations, we will make the following assumptions:
\begin{itemize}
    \item The accuracy, precision and recall are known and reliable for our AI models. We will test in a setting where accuracy, recall and precision have the same values, indicating a balanced models. And we will test values 90\%, 95\% and 99\% for these metrics.
    \item To avoid speculative computation, we will consider that each person living in these metropolitan cities will be tested once, and that the AI algorithms remain consistent in their predictions (i.e.: even if a person were to be analyzed several times, the classification would remain the same).
\end{itemize}

Since it is challenging to determine the exact percentage of criminals within the general population, we have used estimates available online for the four cities, all derived from crime rates. We acknowledge that this approach is not ideal and may not be entirely accurate. A more uniform criminal percentage across all cities might have provided a more consistent comparison. However, we believe that exploring a broader range of settings will actually offer valuable insights into the behavior of AI models across different scenarios and provides us with a better opportunity to study these models.

\begin{table}[!h]
\begin{tabular}{|p{5cm}|c|c|c|c|}
\hline
 & London & Beijing & Hyderabad & New York \\
\hline
\hline
Estimated Population & 9.6M & 21.4M & 11M & 8.4M \\
\hline 
Estimated number of CCTV cameras & 631,627 & 1.31M & $>$600,000 & 70,882 \\
\hline 
Estimated number of camera per 1000 people & 65.76 & 61.21 & 32.4 & 7.35 \\
\hline
Estimated number of daily CCTV exposure by person & 6-300 & 9-279 & 5-148 & 1-34 \\
\hline
Crime rates per 1000 people & 93 & 0.56 & 2.32 & 16 \\
\hline
Estimated percentages of criminals & 1-2\% & 0.1-0.2\% & 0.2-0.5\% & 1-2\% \\
\hline
\hline
Number of misclassified people (90\% accuracy) & 960,000 & 2.1M & 1.1M & 840,000 \\
\hline
Number of false positive cases (90\% precision and recall) & 9,600-19,200 & 2,140-4,280 & 2,200-5,500 & 8,400-16,800 \\
\hline
\hline
Number of misclassified people (95\% accuracy) & 480,000 & 1.05M & 550,000 & 420,000 \\
\hline
Number of false positive cases (95\% precision and recall) & 4,800-9,600 & 1,070-2,140 & 1,100-2,750 & 4,200-8,400 \\
\hline
\hline
Number of misclassified people (99\% accuracy) & 96,000 & 210,000 & 110,000 & 84,000 \\
\hline
Number of false positive cases (99\% precision and recall) & 960-1,920 & 214-428 & 220-550 & 840-1,680 \\
\hline
\end{tabular}
\caption{Conservative estimates statistics of crime prediction failure in 4 large metropolitan cities using 3 different levels of accuracy, precision and recall.}
\label{tab:cctv}
\end{table}

Assuming a population of size $N$ and an AI model for which the accuracy, recall and precision are known. Then, from an estimated percentage of criminals $r$ in our population $N$, computing the number of false positive is done as follows. First, we compute the number of true positive cases:

\begin{equation}
    TP = recall \times r \times N
\end{equation}

Then, we can re-arrange Equation \eqref{eq:prec} which gives us:

\begin{equation}
    FP = \frac{1-precision}{precision} \times TP = \frac{(1-precision) \times (recall \times r \times N)}{precision}
\end{equation}

All results for this simulation are shown in Table \ref{tab:cctv} which also includes contextual information. The results clearly indicate that the number of misclassified people remains extremely high, often exceeding the actual criminality rate in many scenarios. 
Since our paper is focusing on the harm potential of AI methods, the number for false positive cases, which are far from zero, is of particular concern. . It is evident from our results that a large-scale implementation of AI systems based on physiognomy or Lombrosianism would have significant and harmful consequences, with serious implications for people's lives and well-being. With a 95\% precision and recall -which is within the norm for current state-of-the-art methods-, London for instance would potentially feature 4800 to 9600 wrongly convicted people by such AI systems.

Furthermore, the following points are worth noting:
\begin{itemize}
    \item Most algorithms in the literature have an accuracy ranging from 85\% to 95\%, and many of them do not report the precision. We also conservatively assumed that precision, recall and accuracy would have similar values. This is rarely the case.
    \item We also assumed a consistent algorithm would be applied uniformly across all scenarios. If this assumption is incorrect, the actual number of false positive cases could be significantly higher, especially considering the estimated daily exposure to multiple cameras.
\end{itemize}

A major limitation of this simulation is that it does not account for individuals who may be falsely identified multiple times. In other words, the raw numbers presented do not address potential biases affecting different population groups. This issue will be explored in the next section.

\section{The Myth of theory-free inference and unbiased training data}

In an ideal world, one might contend that scientific practices should be value-free in order to minimize harm potential and reduce bias.
However, numerous philosophers and scientists have compellingly argued that Science and the knowledge it produces are shaped by human normative values \cite{longino1990science,zamora2010douglas}. Thus, the notion ``value-free" Science may well be a myth. Indeed, scientific research is always conducted within a broader context, and its value depends on the specific applications it serves and its direct (or indirect) impacts on human lives \cite{popper2013realism}. In essence, any scientific research that serves a purpose can never truly be ``value-free".

In the field of Machine Learning and Artificial Intelligence, a parallel concept to ``value-free" science has emerged in the form of the so-called ``theory-free" models \cite{andrews2023immortal,andrews2024reanimation}. Proponents of \emph{theory-free} models argue that because these models do not rely on specific mechanisms from application fields and are ``data-driven", they would be free from human biases, preconceived judgments, and ontological categories.  We contend that the argument of ``theory-free" AI models is a fallacy, scientific quackery, and far too often serves as a smoke screen to legitimize bigotry through a ``data-driven" pseudo-truth:

First,  to suggest that explanations can be developed without a model or theory undermines the scientific method. The notion that that a large amount of data would remove the need for theory and mechanisms -or worst imply that said theory and mechanism do not exist or can be denied because of empirical evidences \cite{anderson2008end}- is fundamentally flawed. Disregarding theory in applied fields further implies that experts knowledge the historicity of a given field are irrelevant. This is certainly not the case, and field knowledge remains in our opinion the best way to avoid repeating the errors of the past.

Second, the belief that ``data-driven" models are inherently fair and that the patterns and conclusions drawn from large datasets represent undeniable truths is a clear fallacy. Historical proponents of physiognomy and eugenics also subscribed to the idea that good research practices should consists in ``gathering as many facts as possible without any theory or general principle that might prejudice a neutral and objective view of these facts" \cite{jackson2005origins}. This perspective mirrors the ``data-driven" AI methods that have resurrected physiognomy and other pseudosciences under the guise of modern AI. Regardless of the validity of patterns identified by these methods, constructing a model or ideology based solely on correlation, but without understanding causation, is a profound misunderstanding.

Then, the assertion that Machine Learning and Artificial Intelligence are ``theory-free" and ``model-free" and, therefore ``value-free" is incorrect. Although  many AI and DL models do not rely on any specific model, we have seen how they work in section \ref{sec:DL}: They have complicated models, sometimes called ``black-box models", but they have models nonetheless. And while these models indeed appear ``theory-free" because they do not follow explicit preconceived mechanisms, they are far from value free. Indeed, the selection of datasets, features, and variables used in these models is deeply influenced by human values and biases, and can only result in models that are far from ``value-free" \cite{leonelli2019distinguishes}.

Finally, several researchers, philosophers and policy makers argue that the issue of bias in AI systems could be mitigated by using only datasets that have carefully been curated to remove biases and imbalances \cite{mehrabi2021survey,lin2021engineering,obermeyer2019dissecting,doi:10.1073/pnas.1919012117,norori2021addressing}. And it is true that in many fields, such as medicine, biases have shown to be a problem \cite{lin2023improving,}. While we support the use of curated datasets to reduce potential biases and discrimination in AI systems, we firmly believe that achieving perfect fairness and a truly ``value-free" model using only data curation is an unattainable goal for several reasons:
\begin{itemize}
    \item Large Database Requirements: Deep Learning models require vast amounts of data for training. All existing data sources are biased in some way, and there is no such thing as raw unbiased data \cite{mussell2014raw}.
    \item Inherent ideological biases: All methods to re-balance data or remove biases are themselves subject to ideological biases, which can influence the outcomes and effectiveness of these methods.
    \item Limits of removing sensitive features: the idea that balancing data  \cite{wang2019balanced} or removing sensitive features (such as gender, ethnic background, age, etc.) from training data would help is usually wrong: In many cases, the other features are enough for any Deep Learning models to  reconstruct these sensitive attributes as hidden deep features.This approach also hampers explainability \cite{kelly2021challenging}, as biases may persist in ways that are not transparent. Conversely, removing too many key features can significantly impair model performances.
    \item Persistent application bias: Ultimately, and to circle back to the first assertion of this section, we believe that the inherent biases of the application itself contribute to the persistence of biases. As long as an application has potential biases, completely eliminating biases from data sources and features remains unattainable. 
\end{itemize}

\section{Conclusion}

In this paper, we have explored the troubling resurgence of pseudoscientific methodologies within the realm of Artificial Intelligence. In particular, we have discussed how the Deep Learning technology made it easier to hide the pseudoscientific nature of some applied tasks due to their inherent complexity, black-box type model, but also thanks to their seemingly high accuracy. Our analysis further highlights a critical issue: despite their advanced capabilities, these AI systems have often neglected fundamental lessons from statistics, and in particular the principle that correlation does not imply causation.

We have shown how the high performances of these models and reliance on the ``theory-free" ideology made it possible to inadvertently replicate and even exacerbate the errors of past pseudosciences. This includes approaches reminiscent of Lombrosianism and physiognomy, which once justified discriminatory practices through dubious correlations. We have further demonstrated that many state-of-the-art AI models promoting such pseudosciences, by focusing excessively on performance metrics that disregard the harm risks caused by false positive identifications, would be a serious danger if they were to be implemented at large scale.

Our findings suggest that merely addressing biases in training data is insufficient for mitigating the risks posed by these technologies. Instead, a more comprehensive strategy is required. Such strategy should start by a better training of future ML expert to include ethic courses so that they can tell in advance what applications are ethically and morally acceptable or not. Further key elements of the strategy should involve a fundamental reflection on Deep Learning models and ``theory-free" models at large, the systematic use of quality metrics relevant to assess the harm potential of an AI model, and stringent human oversight by field experts. Only through these measures can we hope to curtail the harmful impacts of AI systems and prevent the further promotion of pseudoscientific algorithms under the guise of advanced Deep Learning technology.

As AI continues to evolve and integrate deeper into our society, it is our strong belief that we must heed the lessons from statistics and history. By doing so, we can ensure that these technologies contribute positively to society without perpetuating biases or errors of the past.



\end{document}